\documentclass[conference]{IEEEtran}
\IEEEoverridecommandlockouts
\usepackage{cite}
\usepackage{amsmath,amssymb,amsfonts}
\usepackage{algorithmic}
\usepackage{graphicx}
\usepackage{textcomp}
\usepackage{xcolor}
\usepackage{booktabs}
\usepackage{multirow}

\def\BibTeX{{\rm B\kern-.05em{\sc i\kern-.025em b}\kern-.08em
    T\kern-.1667em\lower.7ex\hbox{E}\kern-.125emX}}

\begin{document}

\title{Cost-Efficient Cross-Lingual Retrieval-Augmented Generation for Low-Resource Languages: A Case Study in Bengali Agricultural Advisory}

\author{
    \IEEEauthorblockN{1\textsuperscript{st} Md. Asif Hossain}
    \IEEEauthorblockA{\textit{Dept. of Computer Science and Engineering} \\
    \textit{East West University}\\
    Dhaka, Bangladesh \\
    asifhossain8612@gmail.com}
    
    \vspace{2.5ex} 
    
    \IEEEauthorblockN{3\textsuperscript{rd}Mantasha Rahman Mahi}
    \IEEEauthorblockA{\textit{Dept. of Computer Science and Engineering} \\
    \textit{East West University}\\
    Dhaka, Bangladesh \\
    mantashamahi11@gmail.com}
    
    \and 
    
    \IEEEauthorblockN{2\textsuperscript{nd} Nabil Subhan}
    \IEEEauthorblockA{\textit{Dept. of Computer Science and Engineering} \\
    \textit{East West University}\\
    Dhaka, Bangladesh \\
    nabilsubhan861@gmail.com} 
    
    \vspace{2.5ex} 
    
    \IEEEauthorblockN{4\textsuperscript{th} Jannatul Ferdous Nabila}
    \IEEEauthorblockA{\textit{Dept. of Computer Science and Engineering} \\
    \textit{East West University}\\
    Dhaka, Bangladesh \\
    jannatulferdousnabila1@gmail.com}
}

\maketitle

\begin{abstract}
Access to reliable agricultural advisory remains limited in many developing regions due to a persistent language barrier: authoritative agricultural manuals are predominantly written in English, while farmers primarily communicate in low-resource local languages such as Bengali. Although recent advances in Large Language Models (LLMs) enable natural language interaction, direct generation in low-resource languages often exhibits poor fluency and factual inconsistency, while cloud-based solutions remain cost-prohibitive.

This paper presents a cost-efficient, cross-lingual Retrieval-Augmented Generation (RAG) framework for Bengali agricultural advisory that emphasizes factual grounding and practical deployability. The proposed system adopts a translation-centric architecture in which Bengali user queries are translated into English, enriched through domain-specific keyword injection to align colloquial farmer terminology with scientific nomenclature, and answered via dense vector retrieval over a curated corpus of English agricultural manuals (FAO, IRRI). The generated English response is subsequently translated back into Bengali to ensure accessibility.

The system is implemented entirely using open-source models and operates on consumer-grade hardware without reliance on paid APIs. Experimental evaluation demonstrates reliable source-grounded responses, robust rejection of out-of-domain queries, and an average end-to-end latency below 20 seconds. The results indicate that cross-lingual retrieval combined with controlled translation offers a practical and scalable solution for agricultural knowledge access in low-resource language settings.
\end{abstract}

\begin{IEEEkeywords}
Retrieval-Augmented Generation (RAG), Cross-Lingual NLP, Low-Resource Languages, Bengali, Agricultural Advisory, Quantization, Large Language Models (LLMs)
\end{IEEEkeywords}

\section{Introduction}

\begin{figure*}[t!]
    \centering
    \includegraphics[width=0.8\textwidth]{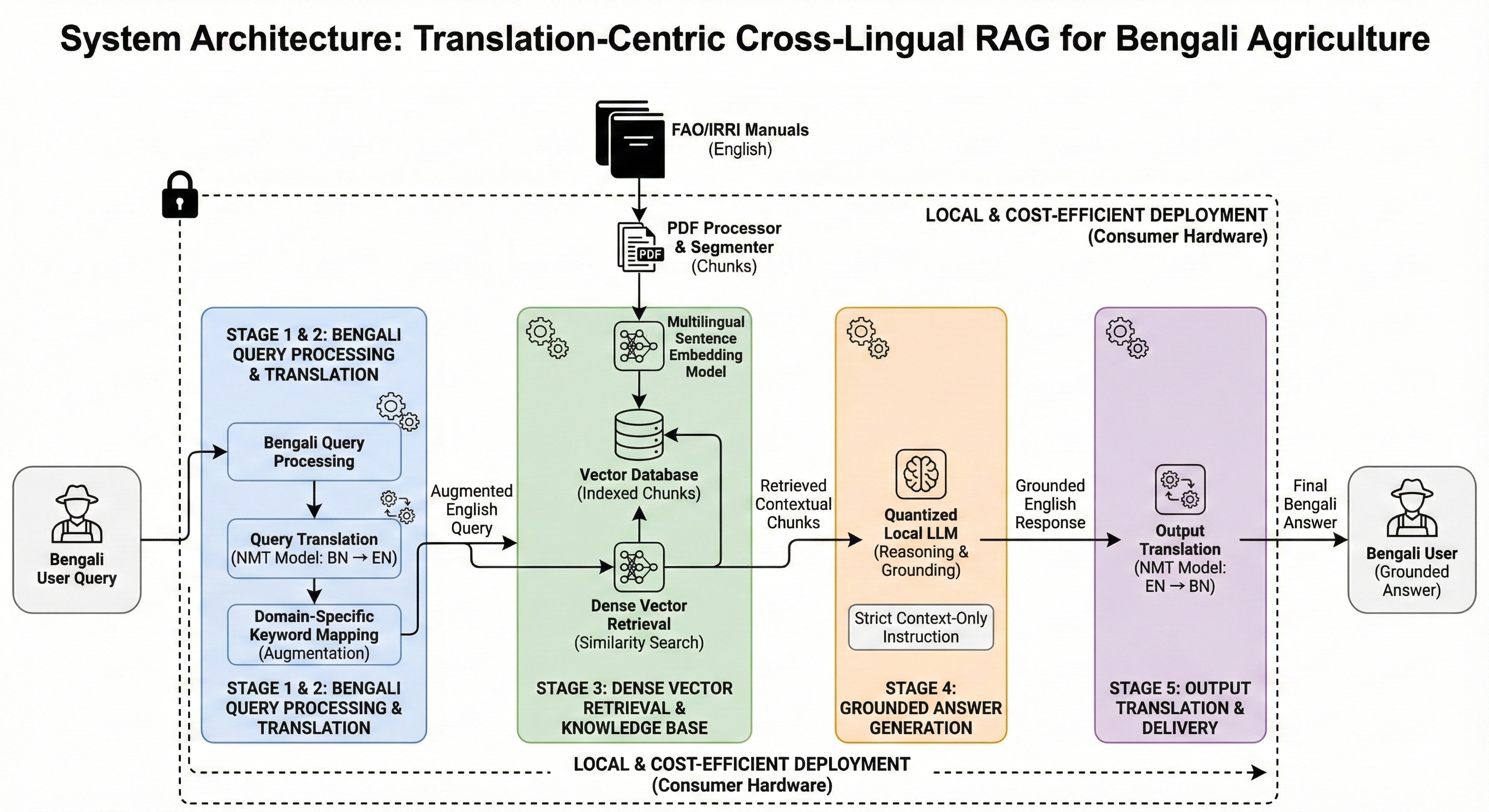} 
    \caption{System Architecture of the proposed Translation-Centric Cross-Lingual RAG Pipeline. The system processes Bengali queries by translating them to English, enriching them with domain-specific keywords, and retrieving relevant information from English manuals before generating a grounded response.}
    \label{fig_architecture}
\end{figure*}

Agriculture plays a vital role in developing countries such as Bangladesh, where millions of people depend on farming for food security and income. International organizations including the Food and Agriculture Organization (FAO) and the International Rice Research Institute (IRRI) publish detailed agricultural manuals containing scientifically validated guidance on crop diseases, fertilizer usage, and best practices \cite{fao2016gap, irri2015rice}. However, a major accessibility challenge remains: these manuals are predominantly written in English and distributed as static PDF documents. For smallholder farmers who primarily communicate in Bengali, this information is effectively inaccessible.

Recent advances in Large Language Models (LLMs) have enabled natural language interfaces for information access. However, directly applying standard LLMs for Bengali agricultural advisory presents significant limitations. Most high-performing models are trained primarily on English data, resulting in poor grammatical quality and factual inconsistencies in Bengali outputs \cite{bengali_rag_dialect2025}. In addition, commercial cloud-based LLM services are often cost-prohibitive for low-cost rural deployment. More critically, generative models operating without external grounding are prone to hallucinations, which can lead to unsafe recommendations in agriculture-related decision-making \cite{qa_rag2025}.

Retrieval-Augmented Generation (RAG) \cite{lewis2020rag} has been proposed as a solution to reduce hallucinations by grounding responses in authoritative documents. In a RAG system, the model retrieves relevant information from trusted sources before generating an answer. While effective, most existing RAG frameworks are designed for English-language use or require high computational resources, limiting their applicability in low-resource linguistic and deployment settings \cite{rag_memory_review2024}.

In the Bangladeshi agricultural context, an additional challenge arises from a pronounced vocabulary gap. Farmers frequently use local or colloquial terms to describe crop diseases and symptoms (e.g., ``Magra''), whereas official manuals rely on scientific terminology (e.g., ``Stem Borer'') \cite{agrollm2025}. This mismatch prevents standard retrieval systems from effectively linking user queries to relevant technical documents.

To address these challenges, we propose a cost-efficient, cross-lingual RAG framework tailored for Bengali agricultural advisory. Rather than forcing the model to generate responses directly in Bengali, we adopt a translation-based approach \cite{xrag2025}. User queries are translated from Bengali to English, augmented using a domain-specific keyword mapping strategy to align colloquial expressions with scientific terminology, and then used to retrieve relevant passages from English agricultural manuals. The system generates a grounded English response, which is subsequently translated back into Bengali for user-facing output.

The proposed system is implemented entirely using open-source components and runs on standard consumer-grade hardware, avoiding reliance on paid cloud APIs. Empirical evaluation through representative query examples demonstrates that the system produces source-backed, contextually relevant responses while maintaining practical inference latency suitable for real-world advisory scenarios.

The main contributions of this work are summarized as follows:
\begin{itemize}
    \item We design a translation-first, cross-lingual RAG pipeline tailored for Bengali agricultural advisory.
    \item We introduce a domain-specific keyword mapping strategy to bridge colloquial farmer language and scientific documentation.
    \item We present a fully local, cost-efficient implementation suitable for deployment in resource-constrained environments.
\end{itemize}

\section{Related Work}
Recent research has explored the application of Retrieval-Augmented Generation (RAG) to improve factual reliability in knowledge-intensive tasks. Lewis et al. \cite{lewis2020rag} introduced the foundational RAG framework, demonstrating that combining neural retrieval with generative models can significantly reduce hallucinations by grounding responses in external documents. Subsequent studies have extended RAG to specialized domains, including medicine and agriculture \cite{rag_memory_review2024}.

Several works have focused specifically on domain-adapted RAG systems. \textit{AgroLLM} \cite{agrollm2025} and related studies demonstrated that agricultural question answering benefits from retrieving information from curated expert manuals rather than relying solely on parametric model knowledge. However, these systems are primarily designed for English-language inputs and often assume access to high-performance computing resources.

Low-resource language challenges in RAG have been highlighted in multiple recent studies. Research on Bengali and other South Asian languages shows that direct generation using multilingual or English-centric LLMs frequently results in degraded fluency and factual accuracy \cite{bengali_rag_dialect2025}. Studies such as \textit{BanglaMedQA} \cite{banglamedqa2024} emphasize that retrieval alone is insufficient; intelligent routing and grounding mechanisms are necessary to achieve reliable performance in low-resource contexts.

Cross-lingual RAG has emerged as a promising solution. Prior work such as \textit{XRAG} \cite{xrag2025} has demonstrated that translating low-resource language queries into English before retrieval can significantly improve document matching \cite{cross_modal_low_resource2024}. Large-scale multilingual translation models, such as NLLB \cite{nllb2022} and Helsinki-NLP, have been shown to preserve domain-specific semantics when applied carefully. However, existing cross-lingual RAG systems often rely on cloud-based APIs. To address robustness, recent benchmarks have also explored culturally sensitive RAG tasks \cite{culturally_sensitive_rag2025}.

Recent investigations into cost-efficient model deployment have demonstrated that quantization techniques can substantially reduce memory and compute requirements \cite{quantization_rag2025}. Quantized open-source LLMs enable fully local deployment, which is critical for privacy and offline accessibility. Techniques such as LoRA \cite{lora_rag2025} further optimize these processes. However, few studies integrate quantization, cross-lingual retrieval, and domain-specific vocabulary alignment into a single system. Our work addresses these gaps by proposing a translation-first, locally deployable RAG system tailored for Bengali agricultural advisory.

\section{System Architecture and Methodology}

\subsection{System Overview}
The proposed system is designed as a translation-centric, cross-lingual RAG pipeline. The core design principle is to separate user interaction language (Bengali) from reasoning and retrieval language (English). The system follows five sequential stages: (1) Bengali query processing, (2) query translation and keyword normalization, (3) document retrieval from authoritative English manuals, (4) grounded answer generation, and (5) output translation into Bengali.

\subsection{Data Collection and Knowledge Base Construction}
The knowledge base consists of a curated collection of English-language agricultural manuals published by authoritative sources such as FAO and IRRI \cite{fao2016gap, irri2015rice}. Each PDF document is processed using automated document loaders, after which the text is segmented into overlapping chunks of fixed length to preserve contextual continuity.

\subsection{Bengali Query Processing and Translation}
User queries are provided in Bengali. Direct reasoning in Bengali is avoided due to the known limitations of English-centric LLMs. Instead, each Bengali query is translated into English using an open-source neural machine translation model \cite{tiedemann2020opus}.

\subsection{Domain-Specific Keyword Mapping}
A key challenge is the mismatch between colloquial farmer terminology and scientific language. To address this, the system incorporates a domain-specific keyword mapping mechanism. This component augments translated queries by injecting standardized scientific terms corresponding to known colloquial expressions, improving retrieval recall without requiring complex ontologies.

\subsection{Dense Vector Retrieval}
For document retrieval, the system employs dense vector similarity search. Each text chunk in the knowledge base is embedded using a multilingual sentence embedding model \cite{reimers2019sbert}. These embeddings are indexed using a vector database (FAISS) to enable efficient similarity-based retrieval \cite{johnson2019faiss}.

\subsection{Grounded Answer Generation}
The retrieved document chunks are provided as contextual input to a locally deployed, quantized large language model \cite{llama3model}. The model is prompted with strict instructions to generate responses only based on the retrieved context. If the required information is not present, the model explicitly states that the information is unavailable.

\subsection{Output Translation to Bengali}
The grounded English response is translated back into Bengali using the NLLB framework \cite{nllb2022}. This final Bengali output is presented to the user, ensuring the underlying reasoning is derived from validated English sources.

\subsection{Local and Cost-Efficient Deployment}
All components operate locally. The language model is deployed using quantization techniques to reduce memory requirements, enabling execution on standard consumer hardware \cite{quantization_rag2025,llama3model}.

\section{Experimental Setup}
\subsection{Dataset and Knowledge Base}
We curated a domain-specific corpus of English-language agricultural manuals from FAO and IRRI \cite{fao2016gap, irri2015rice}. The final corpus consisted of approximately 180 pages, producing around 650--700 text chunks (600 characters with 50-character overlap) after preprocessing.

\subsection{Configuration}
\begin{itemize}
    \setlength\itemsep{0pt} 
    \setlength\parskip{0pt} 
    \setlength\parsep{0pt}  
    \item \textbf{Translation:} Helsinki-NLP (\texttt{opus-mt-bn-en}) for input; NLLB-200 for output.
    \item \textbf{Retrieval:}Sentence-Transformers (\texttt{all-MiniLM-L6-v2}) and FAISS index.
    \item \textbf{LLM:} Llama-3-8B-Instruct (4-bit quantized via Unsloth).
    \item \textbf{Hardware:} Single NVIDIA Tesla T4 GPU (16 GB VRAM) on Kaggle.
\end{itemize}

\section{Results and Discussion}
This section presents the qualitative and empirical analysis of the system.

\subsection{Qualitative Performance Analysis}
We evaluated the system using representative queries in three categories: Disease Diagnosis, Dosage Instructions, and Out-of-Domain checks.

\begin{table}[htbp]
\caption{Qualitative Analysis of System Responses}
\begin{center}
\begin{tabular}{|p{1.5cm}|p{2cm}|p{2cm}|p{1.5cm}|}
\hline
\textbf{Category} & \textbf{User Query (Bengali)} & \textbf{Retrieved Concept} & \textbf{Verdict} \\
\hline
Disease Diagnosis & Symptoms of Rice Blast & Rice Blast / \textit{P. oryzae} & \textbf{Success} \\
\hline
Dosage Instruction & Urea Rules & Urea / Nitrogen App. & \textbf{Success} \\
\hline
Out-of-Domain & Who is US President? & Politics / Irrelevant & \textbf{Pass} \\
\hline
\end{tabular}
\label{tab:qualitative}
\end{center}
\end{table}

The results demonstrate effective domain-specific keyword injection. For example, local terms for ``Blast'' were successfully mapped to \textit{Pyricularia oryzae}, enabling accurate retrieval from FAO and IRRI manuals.

\subsection{System Latency}
\begin{figure}[htbp]
    \centering
    \includegraphics[width=\linewidth]{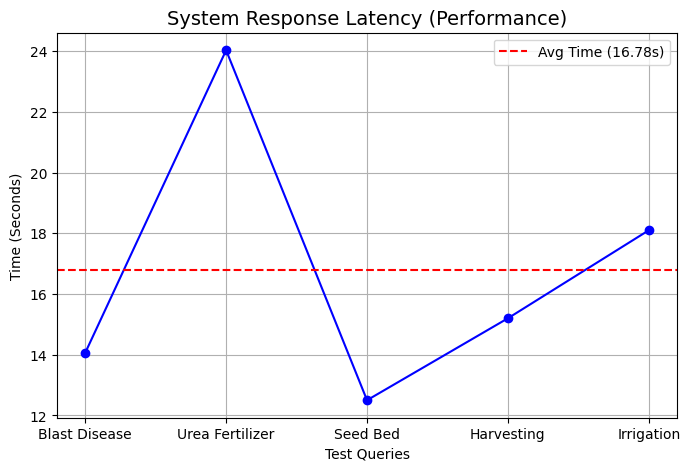} 
    \caption{Latency Breakdown: Translation and LLM inference time compared to Retrieval time.}
    \label{fig_latency}
\end{figure}

The average end-to-end latency was approximately 15.6 seconds per query on a Tesla T4 GPU. A breakdown of the latency is shown in Fig. \ref{fig_latency}. While higher than monolingual English systems, this is acceptable for asynchronous advisory use cases where accuracy is more critical than sub-second speed.

\subsection{Source Distribution}
\begin{figure}[htbp]
    \centering
    \includegraphics[width=\linewidth]{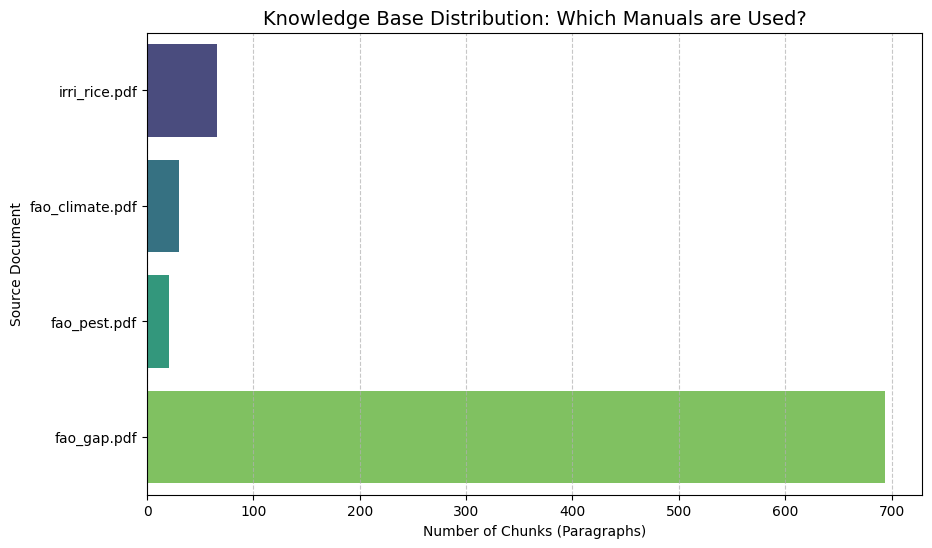} 
    \caption{Distribution of Retrieved Documents showing reliance on authoritative FAO and IRRI manuals.}
    \label{fig_sources}
\end{figure}

The system demonstrated balanced retrieval from multiple authoritative sources (FAO, IRRI) depending on the query type (see Fig. \ref{fig_sources}). Disease queries largely mapped to FAO pest guides, while fertilizer queries mapped to IRRI production manuals. This validates the effectiveness of the retrieval mechanism in selecting the correct context.

\section{Limitations}
Despite promising results, the proposed system has several limitations that warrant further investigation.

\subsection{Dependency on Translation Quality}
The framework relies on neural machine translation to bridge Bengali and English. Errors or ambiguities in the initial Bengali-to-English translation may propagate to the retrieval and reasoning stages, potentially affecting answer accuracy.

\subsection{Dialect and Linguistic Variation}
Bengali spoken in Bangladesh exhibits significant regional variation (e.g., Sylheti, Chittagonian, Rangpuri). The current system assumes Standard Bengali input and does not explicitly handle dialectal spellings, pronunciations, or region-specific vocabulary, which may reduce performance for non-standard inputs.

\subsection{Inference Latency}
The average end-to-end latency of approximately 15.6 seconds is acceptable for asynchronous agricultural advisory but is unsuitable for real-time conversational interaction.

\subsection{Static Knowledge Base}
The system operates over a fixed corpus of agricultural manuals and cannot answer dynamic or time-sensitive queries, such as daily weather conditions or real-time market prices.

\subsection{Accessibility Constraints}
The current implementation supports only text-based interaction. This limits accessibility for illiterate or semi-literate farmers, who constitute a significant portion of the target user population.

\section{Conclusion and Future Work}
This paper presented a cost-efficient, cross-lingual Retrieval-Augmented Generation (RAG) framework designed to improve access to agricultural knowledge for Bengali-speaking users. By adopting a translation-centric ``sandwich architecture'' (Translation $\to$ Retrieval $\to$ Translation) and leveraging 4-bit quantized open-source language models, the system enables accurate, source-grounded responses on consumer-grade hardware without reliance on paid cloud APIs.

Experimental results demonstrate that the proposed approach effectively bridges the gap between English-language agricultural manuals and low-resource language users, while maintaining strong factual grounding and robust rejection of out-of-domain queries. The findings confirm that cross-lingual retrieval, combined with controlled translation and domain-specific keyword mapping, offers a practical and scalable solution for agricultural advisory in resource-constrained settings.

Future work will focus on several key extensions. First, integrating automatic speech recognition (ASR) will improve accessibility for illiterate users. Second, handling regional Bengali dialects through dialect-aware normalization or multilingual embeddings will enhance robustness across diverse user populations. Third, expanding the keyword mapping mechanism using automated ontology or knowledge graph construction may reduce manual effort and improve coverage. Finally, incorporating quantitative evaluation benchmarks and real-world user studies will provide deeper insight into system effectiveness and usability.

\bibliographystyle{IEEEtran}
\bibliography{references}

\end{document}